# Development of a Knowledge Graph Embeddings Model for Pain

Jaya Chaturvedi, MSc[1], Tao Wang, PhD[1], Sumithra Velupillai, PhD[1], Robert Stewart, MD[1,2], Angus Roberts, PhD[1,2]

[1]Institute of Psychiatry, Psychology and Neurosciences, King's College London, London, United Kingdom; [2]South London and Maudsley NHS Foundation Trust, London, United Kingdom

**Abstract**

*Pain is a complex concept that can interconnect with other concepts such as a disorder that might cause pain, a medication that might relieve pain, and so on. To fully understand the context of pain experienced by either an individual or across a population, we may need to examine all concepts related to pain and the relationships between them. This is especially useful when modeling pain that has been recorded in electronic health records.*

*Knowledge graphs represent concepts and their relations by an interlinked network, enabling semantic and context-based reasoning in a computationally tractable form. These graphs can, however, be too large for efficient computation. Knowledge graph embeddings help to resolve this by representing the graphs in a low-dimensional vector space. These embeddings can then be used in various downstream tasks such as classification and link prediction.*

*The various relations associated with pain which are required to construct such a knowledge graph can be obtained from external medical knowledge bases such as SNOMED CT, a hierarchical systematic nomenclature of medical terms. A knowledge graph built in this way could be further enriched with real-world examples of pain and its relations extracted from electronic health records. This paper describes the construction of such knowledge graph embedding models of pain concepts, extracted from the unstructured text of mental health electronic health records, combined with external knowledge created from relations described in SNOMED CT, and their evaluation on a subject-object link prediction task. The performance of the models was compared with other baseline models.*

**Introduction**

Pain is a global health problem and is estimated to affect 1 in 5 adults worldwide (1). Pain is a massive burden on society in terms of costs related to medical care as well as loss of productivity (2). A committee reviewing the public health significance of pain in the United States found that the total cost to society was greater than that estimated for heart disease, cancer or diabetes (3). People who experience chronic pain are more likely to develop emotional distress, which can create muscle tensions and increase pain. There is a known intersection between pain and mental health disorders, such as pain and depression (4), bipolar and psychotic disorders (5). Consequently, the impact of pain on mental health and quality of life is an active area of research. Pain is a common reason for people to seek medical attention (6), and is therefore widely described in electronic health records (EHRs). In mental health EHRs, patients' experiences of pain are often recorded as free-text. EHRs have therefore become a valuable resource in the research of pain (7,8). There is substantial evidence to support an overlap between pain and mental health (9,10). Compared to physical health conditions, more contextual information is generally required and therefore recorded about pain for patients with mental health conditions, making the clinical text within mental health EHRs a good source of such information.

Knowledge graphs (KGs) are large networks which allow for the representation of entities/concepts, along with their semantic types and relations to other entities as graphs (11) . They have emerged as an efficient method of representing data as a heterogeneous graph, facilitating the visualization of and reasoning over complex data and its interconnected relationships, which further help reveal any hidden patterns and deduce new knowledge (12). A KG typically consists of a set of fact triples, referred to as subject-predicate-object triples, or nodes and edges, or head-relation-tail triples (13). For example, in a triple such as <paracetamol, relief, pain>, paracetamol is the subject/node/head, relief is the predicate/edge/relation, and pain is the object/node/tail. In this paper, we will use the terminology subject, predicate and object. KGs can be huge, making them impractical and computationally expensive to use. This issue is resolved by using KG embeddings i.e., low dimensional representations in a vector space (14). Knowledge graph embeddings (KGEs) also assist in further enriching the data by representing the semantics of domain knowledge within the KGs

(15). KGE models learn embeddings of the entities and relations based on scoring functions that predict the probability that a given triple is a fact, i.e., higher scores indicate a true triple or more likely to be factually correct. These scoring functions combine the embeddings of the triples using different intuitions. The two models used in this work are ComplEx (16) and TransE (17), which are described in more detail in the Methods section. KGE methods have been applied to various biomedical use cases where data is linked to relevant ontologies and terminologies to predict relations (18), understand gene to phenotype associations (19), and predict disease comorbidity (20). The multidimensional nature of pain (21) makes it a good use case for application of such KGE methods. Other EHR-based use cases include patient stratification and drug identification (22), and disease relation extraction (23).

This paper describes the development of KGE models of pain incorporating both pain concepts found within a mental health EHR database, and external knowledge about these concepts from a knowledge base, SNOMED CT (24) (detailed in the Methods section), for use in research on the relationships between mental health, pain, and physical multimorbidities. Whilst it is common to build KGE models from knowledge bases, we have also incorporated information from the EHR, hypothesizing that the addition of real-world language context will improve performance in downstream tasks on EHR text. Three models were constructed by varying the features that were included in the embeddings. The models were evaluated using a link prediction task, and comparisons made between these models. They were also compared with other biomedical and non-biomedical benchmark models that were publicly available. Existing pain research is limited to the use of structured codes in combination with some clinical text from EHRs (7,8) or patient-focused questionnaires and interviews (2,6). There is limited research utilizing clinical text from within EHRs combined with external knowledge bases (23,25). To the best of our knowledge, this is the first time such KGE models have been developed for pain research. The models, and scripts used to develop them, are publicly available[1] and could be adapted for use in other areas of medicine.

**Methods**

*Data Collection*
EHR text was extracted from an anonymized version of a large mental health EHR from The South London and Maudsley NHS Foundation Trust (SLaM) through its Clinical Record Interactive Search (CRIS) data platform (26). The infrastructure of CRIS has been described in detail with an overview of the cohort profile (26). CRIS is comprised of over 30 million documents and over 500,000 patient records (26), averaging about 90 documents per patient (27).

SNOMED CT (24) is one of the most commonly used medical knowledge bases in healthcare, and so has been used in various KGE models (15,28). The formal and hierarchical structure of SNOMED CT facilitates the classification of data into different taxonomic categories which combine various clinical concepts such as diseases and medications (29). These add a level of semantics to clinical data by providing reference to different concepts and the relationships that exist between them, thereby enabling logical reasoning. In combination with natural language processing (NLP), such structured knowledge can help disambiguate concepts mentioned within the unstructured clinical notes of EHRs and produce more meaningful results (24). One advantage of SNOMED CT over other biomedical terminologies is that it is designed as a compositional, post-coordinated system. This compositional design ensures that when SNOMED CT is used in different systems and contexts, it will still produce the same conceptual and computational meaning for concepts (24). It could therefore be a valuable resource for NLP on clinical data.

As shown in Figure 1, a set of pain keywords were derived from a pain lexicon, development of which is described in (30). A SQL query was run to extract free text documents from the CRIS database (no time or diagnosis filters were applied) that contained any of these pain keywords. Keyword searches can often lead to noise. To refine this, these documents were loaded onto a medical concept annotation tool, MedCAT (31) which was used to pre-annotate all the pain concepts within the documents, linking each concept to a unique SNOMED CT ID (SCTID). Three medical student annotators manually reviewed these pre-annotations and marked them as relevant mentions of physical pain or not. The annotation guidelines that were used by these students are publicly available[2]. In addition to this, SNOMED CT was used to generate the parent and child nodes of every term in the lexicon. This is described in the "Relations" section.

---

[1] https://github.com/jayachaturvedi/pain_kge_model
[2] https://github.com/jayachaturvedi/pain_in_mental_health/blob/main/Annotation%20Guidelines%20-%20Pain%20-%20for%20github.pdf

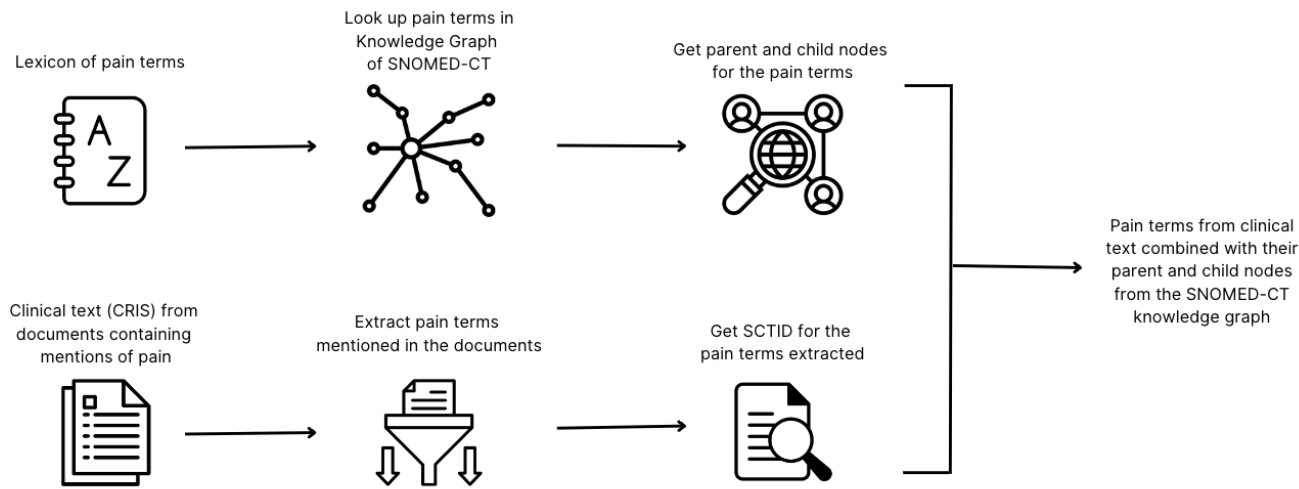

**Figure 1.** Creation of dataset for building KGE models

*Ethics and Data Access*
The source clinical data are accessed through SLaM, the data custodian. Within a customized information governance framework, the Maudsley CRIS platform provides access to anonymized data derived from SLaM's electronic medical records. These data can only be accessed by authorized individuals from within a secure firewall (data cannot be sent elsewhere)[3]. Ethical approval to use the data for research was granted by Oxford C Research Ethics Committee, reference 18/SC/0372.

*Relation Extraction*
A knowledge graph of SNOMED CT was developed using Clinical Knowledge Graph (CKG (32)), which was implemented in the Neo4J graph database format (33). CKG contains 10 different ontologies, including SNOMED CT, and therefore contains every SNOMED CT concept and their parent/child nodes. A query written using Neo4J's Cypher query language was run on CKG to extract the first-order parent and child nodes for all the pain keywords derived from the lexicon. For example, the concept "abdominal pain" can have various relations as shown in Figure 2.

---

[3] Please contact cris.administrator@slam.nhs.uk for more information.

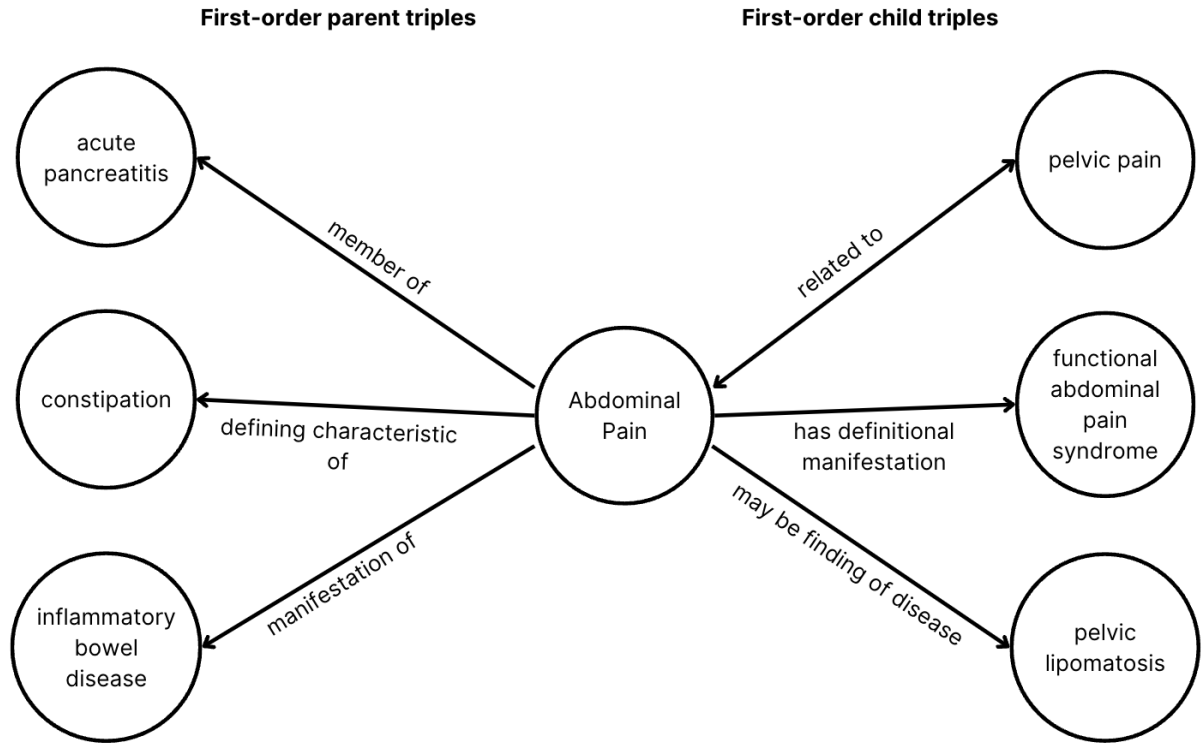

**Figure 2.** An example of first-order parent and child triples for the concept "abdominal pain"

*Knowledge Graph Embedding*
Python version 3.7.16 and the AmpliGraph 1.4 library (34) were used to develop KGE models using the triples generated from CKG. Two commonly used models are ComplEx (16), which uses tensor factorization (a three-way tensor is defined in the form of *n x n x m* where n is the number of entities (subject and object) and m is the number of relations (predicates) - the embeddings are calculated by factorizing this tensor), and TransE (17), which relies on distance (the relationship between subject and object is interpreted as a translation vector so that the embedded entities connected by a relation have a short distance i.e. distance-based functions in the Euclidean space). ComplEx was used since it is considered better at representing multi-dimensional data and preserving asymmetry between concepts such as those defined in biomedical ontologies (35). This was compared to TransE (17) which is commonly used as a benchmark. These two models were chosen because each of them has strengths that may be advantageous in the EHR setting. TransE models asymmetry, inversion and composition, the latter being most useful for SNOMED CT which is inherently compositional in nature. However, TransE lacks the ability to model symmetry, and one-to-many relations. ComplEx models symmetry, asymmetry, inversion, and one-to-many relations. However, it lacks the capacity to capture composition (36).

Three variations of the KGE models were constructed, each variation included both ComplEx and TransE:

Variation 1: The triples of the pain keywords from the lexicon were used in the development of these models. This variation does not include any EHR data.

Variation 2: The triples of pain keywords from the lexicon were combined with the pain concepts within the sentences from CRIS data to form the final dataset for development of these models. This was to ensure incorporation of pain concepts mentioned within CRIS data, but either not in the lexicon, or in the lexicon as variants. For example, "response to pain" and "on examination - painful ear" are concepts found within the sentences but referred to as "pain" and "ear pain" within the lexicon.

Variation 3: In addition to the data used in variation 2, the embeddings of the sentences that contained the pain concepts within CRIS were included in the development of these models. This was to capture the context of the sentences that contained the pain concepts, and makes use of the fact that both models, ComplEx and TransE, represent the triples, pain concepts, and sentences in a shared continuous vector space. This allows for meaningful calculations and enables the models to be used for tasks like link prediction efficiently. In order to represent sentences in the same embedding space, averaging or pooling of the embeddings of the individual words is undertaken, which results in a single vector representing the entire sentence. This shared embedding space between the triples, pain concepts and sentences help capture the semantic relationships between them.

*Link prediction*
The data for each variation was randomly split into training and test sets in the proportion of 80:20. The training set was used to build the models, and the test sets were used for evaluation of the models on a link prediction task i.e., the model is given the subject-predicate and asked to predict the object, and vice versa. The link prediction task is conducted purely for evaluation of performance of the models. Default Ampligraph parameters were used (listed in Table 1). Since the data used for the 3 variations are inherently different, it was not feasible to use the same data for training all the variations, although the triples from variation 1 are common to all 3 variations. To ensure fair comparisons of these variations, a consistent method was used for splitting the date into 80:20 proportions for each variation. 10-fold cross validation was also conducted on the 3 variations, to compensate for the differences in the datasets and make the evaluation more robust.

**Table 1**. KGE model parameters

| Model | Parameters |
| --- | --- |
| ComplEx | [1]batches_count: 100<br>seed: 555<br>epochs: 10<br>[2]k: 150<br>[3]eta: 10<br>loss: 'multiclass_nll' |
| TransE | [1]batches_count: 100<br>seed: 555<br>epochs: 10<br>[2]k: 150<br>[3]eta: 10<br>loss: 'pairwise' |

[1]batches_count: number of batches in which the training set is split during training
[2]k: dimensionality of the embedding space
[3]eta: number of negative, or false triples, generated for each positive, or true triple, during training

For the link prediction task, the metrics will be reported on Mean Reciprocal Rank (MRR) and Hits@N, which are two popular metrics for this type of evaluation task. The ranks in MRR indicate the rank at which the test set triple was found when performing link prediction using the models. Mean Reciprocal Rank (MRR) is a measure of how well a KGE model can predict the missing link (either the subject, the object, or the relation) based on the embeddings learned by the model. For example, given a subject (Ex: "abdominal pain"), and relation (Ex: "may be finding of disease"), the KGE model predicts the rank of each possible object. The reciprocal rank is then calculated for the correct object. In our example, if the model ranked the correct answer (Ex: "pelvic lipomatosis") as 1st choice, the reciprocal rank would be 1, whereas if it were ranked as 2nd choice, the reciprocal rank would be 0.5. A higher MRR indicates that the model is more accurate at finding the correct relationship. Hits@N computes how many elements of a vector of rankings was in the top n positions. Hits@N measures the percentage of correctly predicted entities in the top N ranked results. Hits@1 measures the percentage of correctly predicted entities in the top 1 ranked result, and Hits@10 measures the percentage of correctly predicted entities in the top 10 ranked results. The higher the Hits@N score, the better the model is at predicting missing links in the knowledge graph. During link prediction, the original triples in the knowledge graph are corrupted to form negative examples. For example, for a given positive triple (head, relation, tail) in the knowledge graph, negative examples are created by replacing either the head or the tail entity with some randomly chosen entities. These corrupted triples serve as distractors and are used to evaluate the model's performance. For link predictions, the metric of choice is generally Hits@N since it focuses on the model's ability to rank the correct entity within the top N positions. MRR emphasizes the overall performance while Hits@N emphasizes more on results in top ranking However, it is better to consider multiple metrics, such as MRR in combination with Hits@N, so that we can get a comprehensive understanding of the model's performance.

The two models developed here (ComplEx and TransE) were compared to biomedical (trained on SNOMED CT (15)) and non-biomedical benchmarks (trained on FreeBase (37)) that are commonly used for such tasks.

**Results**

*Data Statistics*
The pain concepts from the lexicon were used to generate triples from the SNOMED CT knowledge graph within CKG (total of 15,336 triples). A portion of these (training set: 80%) were used to generate the KGE models in variation 1. 5,644 sentences from the CRIS data were identified as containing a total of 206 unique pain concepts. These were merged (triples from CKG and the 206 pain concepts from CRIS) to form the final dataset for building the KGE models, 80% of which was used in building the models in variation 2. 80% of the 5,644 sentences were included in the dataset used to build the models in variation 3.

The most frequent triple in our data was *"pain"-" may be treated by"-"aspirin".* The top 5 subjects and predicates are listed in Table 2.

**Table 2.** Top 5 subjects and predicates (Objects not included because the frequency of each was very small (<1%))

| Top 5 | From pain lexicon | | From CRIS data | |
|---|---|---|---|---|
| **Subject** | Pain | 15% | Pain | 42% |
| | Headache | 6% | Chest pain | 6% |
| | Abdominal pain | 6% | Abdominal pain | 4% |
| | Rheumatoid Arthritis | 5% | Headache | 4% |
| | Spasm | 4% | Sore sensation quality | 3% |
| **Predicate** | inverse is a | 25% | may be treated by | 36% |
| | may be treated by | 23% | may be finding of disease | 18% |
| | may be finding of disease | 10% | may be prevented by | 16% |
| | classifies | 6% | inverse is a | 8% |
| | may be prevented by | 5% | classifies | 2% |

*Results of link prediction*
The performance metrics for the link prediction task of the KGE models are given in Table 3. The variation that included pain concepts as well as sentence embeddings from the EHR data (variation 3) performed best, and overall, the ComplEx model performed better than TransE in all instances, with an MRR of 0.88.

**Table 3.** Performance metrics of the two models (ComplEx and TransE) for the three variations, compared to biomedical benchmarks that were trained on SNOMED CT (15) and non-biomedical benchmarks trained on FreeBase (37)

| Models | | Performance Metrics | | |
|---|---|---|---|---|
| | | MRR | Hits@10 | Hits@1 |
| **Non-biomedical benchmark** | **ComplEx** | 0.32 | 0.50 | 0.35 |
| | **TransE** | **0.31** | **0.50** | **0.35** |
| **Biomedical benchmark** | **ComplEx** | 0.46 | 0.65 | 0.36 |
| | **TransE** | 0.34 | 0.59 | 0.21 |
| **Pain KGE without EHR data (Variation 1)** | **ComplEx** | 0.15 | 0.27 | 0.11 |
| | **TransE** | 0.18 | 0.33 | 0.12 |
| **Pain KGE with pain concepts from EHR data (Variation 2)** | **ComplEx** | 0.79 | 0.86 | 0.74 |
| | **TransE** | 0.30 | 0.48 | 0.20 |
| **Pain KGE with pain concepts and sentences from EHR data (Variation 3)** | **ComplEx** | **0.83** | **0.87** | **0.80** |
| | **TransE** | 0.29 | 0.41 | 0.23 |

**Discussion**

This paper describes the development of KGE models with and without utilizing EHR data about pain from a mental health EHR database, combined with external knowledge from SNOMED CT. A link prediction task was used to evaluate the performance of the different models and variations and will not be used in any clinical tasks within this project. The metrics used to evaluate the link prediction, MRR and Hits@N, provide insights into the model's ability to rank true relationships higher than false ones, thereby indicating that the model has learnt the required embeddings that would capture the semantic relationships between the entities. This would in turn be useful and transferable to classification tasks. However, additional evaluations will be carried out on the classification tasks as well, by utilizing the relevant metrics for classification such as precision, recall and F1-score. The ComplEx model performed better than TransE in most variations. This could be because ComplEx has the ability to capture nuanced relationships between entities and relations by representing them as complex vectors. TransE, on the other hand, uses simple vectors. ComplEx performed best in variation 3, which incorporated sentences from the EHR in addition to the pain concepts. Incorporation of more data into the construction of the KGE model meant more relationships between entities, especially of the one-to-many nature, which is a strength for ComplEx-based models. They are able to model multiple relations between entities, while TransE was designed to handle one-to-one relations. The addition of sentence embeddings into the models from variation 3 would also have meant more features to learn from, and therefore better performance. Another strength for ComplEx is its ability to handle noisy data, which EHR data is renowned for. The use of simple vectors in TransE means it is impacted by noise in data. TransE performed best on the non-biomedical benchmark trained on FreeBase. This could be because such data is not as noisy as EHR data. Overall, these two models performed better on link prediction when compared to biomedical and non-biomedical benchmarks. The dataset used for generation of the KGE models in this work is quite small, and specific to pain concepts, which could be why it performed better than the larger biomedical and non-biomedical benchmarks.

**Conclusions**

The ambiguous nature of pain and the complexity of how it is described within text highlights the need for additional information from external knowledge bases to supplement the data available with EHRs, in combination with contextual information from the sentences that contain information about pain. While recent literature has also incorporated such contextual information in their work, they lack the advantage of leveraging the compositional nature of a knowledge base such as SNOMED CT, and instead rely on sources such as ICD-9 (38), DrugBank (29) or niche databases such as the traditional Chinese medicine knowledge base (25).

As part of future work, the ComplEx KGE model built in variation 3 will be used in a downstream binary sentence classification task, to classify sentences as relevant to pain or not. Description of pain in mental health records is mostly restricted to the unstructured free-text of the records. By developing a method to extract information about pain from text, we are able to use that information in studies of pain in the context of mental health. It will allow for better understanding of whether patients with certain mental health disorders report more pain. This could potentially help in early detection of such pain, thereby improving patient outcomes that could have deteriorated due to long durations of untreated pain symptoms (39). Output from this will then be used to explore associations between pain and mental health, and comorbid pain as a predictor of adverse outcomes for people with mental disorders. We have shown that the KGE models that combined information from structured knowledge and real-world textual data from EHRs performed best, which shows potential in performing better at downstream tasks that classically only use EHR data. These results will be compared to those of classifiers built for the same use-case without the incorporation of any external knowledge. While the pipeline to use the classifiers in combination with the KGE model will be more complex due to the need for added information, such as triples for all the pain concepts mentioned within the clinical notes, the benefit of better performance will ensure more accurate classification, and therefore better quality of pain information extraction, which will in turn feed back into better pain research for patient care and pain management. All code used to generate the triples using CKG, the Python code for building and evaluation of the KGE models, and the models themselves, are openly available on GitHub[4].

**Funding and Acknowledgements**

AR was part-funded by Health Data Research UK, an initiative funded by UK Research and Innovation, Department of Health and Social Care (England) and the devolved administrations, and leading medical research charities. RS and AR are part-funded by the National Institute for Health Research (NIHR) Biomedical Research Centre at South London and Maudsley NHS Foundation Trust and King's College London. RS is additionally part-funded by the National Institute for Health Research (NIHR) Applied Research Collaboration South London (NIHR ARC South London) at King's College Hospital NHS Foundation Trust, and by the DATAMIND HDR UK Mental Health Data Hub (MRC grant MR/W014386). AR and RS were additionally part-funded by the UK Prevention Research Partnership (Violence, Health and Society; MR-VO49879/1), an initiative funded by UK Research and Innovation Councils, the Department of Health and Social Care (England) and the UK devolved administrations, and leading health research charities. JC was supported by the KCL funded Centre for Doctoral Training (CDT) in Data-Driven Health. TW was supported by the Maudsley Charity and an Early Career Research Award from IoPPN. The funders had no role in study design, data collection and analysis, decision to publish, or preparation of the manuscript.

---
[4] https://github.com/jayachaturvedi/pain_kge_model